\documentclass[runningheads]{llncs}
\usepackage{graphicx}

\usepackage[dvipsnames]{xcolor}
\usepackage{tabularx}
\usepackage{multirow}
\usepackage{subcaption}

\usepackage{amssymb}

\begin{document}
\title{Category Aware Explainable Conversational Recommendation}

\author{Nikolaos Kondylidis\inst{1}\orcidID{0000-0003-4304-564X} \and
Jie Zou\inst{2}\orcidID{0000-0003-0911-8542} \and
Evangelos Kanoulas\inst{2} \orcidID{0000-0002-8312-0694}}

\authorrunning{N. Kondylidis et al.}

\institute{Vrije Universiteit Amsterdam, Amsterdam, Netherlands \email{n.kondylidis@vu.nl} \and
University of Amsterdam, Amsterdam, Netherlands\\
\email{\{j.zou,e.kanoulas\}@uva.nl}}

\maketitle      

\begin{abstract}

Most conversational recommendation approaches are either not explainable, or they require external user's knowledge for explaining or their explanations cannot be applied in real time due to computational limitations.
In this work, we present a real time category based conversational recommendation approach, which can provide concise explanations without prior user knowledge being required.
We first perform an explainable user model in the form of preferences over the items' categories, and then use the category preferences to recommend items.
The user model is performed by applying a BERT-based neural architecture on the conversation. Then, we translate the user model into item recommendation scores using a Feed Forward Network.
User preferences during the conversation in our approach are represented by category vectors which are directly interpretable.
The experimental results on the real conversational recommendation dataset ReDial \cite{ReDial} demonstrate comparable performance to the state-of-the-art, while our approach is explainable. We also show the potential power of our framework by involving an oracle setting of category preference prediction.

\end{abstract}

\section{Introduction}
Research on conversational recommendation is known to be of high importance\cite{conv_rec_importance_1}. Also, explainable recommendation attracts increasing attention with the popularity of black-box, and inscrutable neural models. Explanations allow users to further trust the recommender, make faster and better decisions and increase their satisfaction \cite{fashion_comment_generation,visual_fusion,explicit_factor_models}. In this work we investigate the explainable conversational recommendation.

We argue that current explainable item recommendation approaches cannot be extended to the conversational setting for cold-start users because they either require external knowledge (e.g. user reviews, social media content, user's interests) \cite{MMALFM,viewpoint_regr,towards_graph,DARIA_SARAH,the_FacT,openionated,explicit_factor_models,EXPLORE} or they cannot be applied in real time (separate execution required for each candidate item) \cite{expl_conv_rec,fashion_comment_generation,visual_fusion,aspera}. Moreover, those works usually are not able to understand natural language conversations. While there is one explainable conversational recommendation study that can understand natural language in a conversational setting, it still has the same two aforementioned limitations \cite{expl_conv_rec}.
Furthermore, studies where review datasets are synthetically translated into conversational recommendation datasets, are explicitly asking user's opinions over items' aspects making them arguably explainable and will not be compared with this work \cite{Christakopoulou,CRS,SAUR,zou2020towardsb}.
Although some studies work on real conversational recommendation datasets, they do not have explainable properties \cite{real_time_conv_rec,KBRD,ReDial,conv_travel}. Summarizing, current studies cannot be applied in a real time cold-start conversational setting while providing explanations for their recommendations.

Our premise is that users make choices on the basis of their preferences for item properties. In the conversation, it is common that users express their sentiments on item properties using natural language. Therefore, if we can extract sentiments for item properties from the conversation and recommend items accordingly, we can then explicitly manipulate the recommendation and also use the sentiments over item properties as an explanation. In this work we focus on improving the item recommendation engine of conversational recommendation systems while providing explanations.
Our proposed approach performs item recommendation by (i) predicting user's preference for each of the items' categories (properties) and (ii) using this prediction as a base to recommend items. For the former part we apply a BERT-based text comprehension model. Regarding the latter part, we use the category preference prediction as the input of a Feed Forward Network to generate item recommendation scores. 
Our contribution in this study is a recommendation framework that
(i) performs explainable user modeling for recommendation
(ii) does not require external knowledge for users (iii) can be applied in real time and (iv) has comparable performance for recommendation accuracy to the state-of-the-art, and can be further improved by a perfect category preferences prediction.
This is the first recommendation work applied in a real time cold-start conversational setting while providing explanations for recommendations, to the best of our knowledge.

\section{Related Work}

In this section, we will briefly review studies performed on conversational recommendation. Studies performed on synthetic conversations generated from the corresponding item related documents (e.g. conversation construction on the basis of entities \cite{zou2020towardsb,zou2019learning,zou2020towards,zou2018technology} or aspects\cite{SAUR}), which allows for only one target item throughout the complete conversation \cite{Christakopoulou,zou2020towardsb,zou2019learning}, are therefore excluded for the comparison. We argue that this assumption does not necessarily hold under a realistic conversational recommendation setting, due to the fact that the user might prefer having several options. To the best of our knowledge, there has not been any study on explainable conversational recommendation applied on real natural language conversations. Please note that the work of \cite{expl_conv_rec} is applied on synthetic conversations (item reviews), it cannot be applied in real time (requires one execution for each item in the collection) and needs prior user knowledge (item reviews).
There are four relevant studies based on real natural language conversations, while non of them has explainable properties. The first three (\cite{KBRD,ReDial,conv_travel}), aim to develop recommender systems that participate in the conversation. The fourth study \cite{real_time_social_rec}) only observes a conversation between two users and recommends advertisements accordingly.

The creators of the \textbf{ReDial} \cite{ReDial} applied three modules in order to build a conversational recommender system: (1) an item recommender, (2) a targeted sentiment analysis module and (3) a natural language response generator module. The recommender is an autoencoder that applies collaborative filtering technique in order to predict movie ratings, following the approach of \cite{autorec}. The recommender follows a black-box setting, not having any explainable properties. Sentiment analysis on mentioned movies is being performed with the encoder part of a Hierarchical Recurrent Encoder Decoder (HRED)~\cite{HRED}. For text comprehension the authors use the encoder part of the HRED and the complete HRED for natural language generation. A switch mechanism \cite{Switch} is being applied on the hidden activations of the HRED. This mechanism decides if the next token to be produced should be a word or a movie. The switch probability value affects the word probabilities and the movie ratings, which are all concatenated together to form output probability distribution over words and movies. The first two modules are being pretrained separately. Additionally, the encoder of the applied HREDs is modified to take general purpose sentence (GenSen) representations from a bidirectional GRU. Pretrained GenSen representations are being used~\cite{pretrained_gensen_representations}, due to the small amount of text in the dataset.

The next two studies \cite{KBRD,conv_travel}, rely on graph information and direct entity relations. The first of them is the only additional study developed on the ReDial \cite{KBRD} dataset. In that study, the text generation model is replaced with a transformer \cite{transformers} but the main difference lies on the recommender model. External knowledge in the form of graph is being introduced from DBpedia \cite{dbpedia}. Graph nodes can either be items entities or non-item entities (e.g. the property "science fiction film"). \cite{KBRD} initially identify items in the conversation through name matching and then apply entity linking \cite{KBRD_entity_linking} in order to expand the matches between the dialogue and the graph by matching non-item entities. A non-item entity is for example \textit{Science\_fiction\_film}\footnote{http://dbpedia.org/resource/Science\_fiction\_film} which can be matched with the sentence "I like science fiction movies" \cite{KBRD}. Relational Graph Convolutional Network is being applied in order to recommend items based on entities that appear in a dialogue. In the second study \cite{conv_travel}, the conversational recommender system proposes venues to travelers. The recommendation in this case is based on three aspects. Firstly, a topic is being predicted given the ongoing conversation. Secondly, conversation and venue similarity is being calculated on a textual level, based on venue's textual description. Thirdly, a graph is formed based on venue information, and a graph convolution network is being applied for improving venue relations, given their descriptions and details.

Last but not least, the conversational recommender system proposed by \cite{real_time_social_rec} has very similar intuition to our approach. Their task is to provide a real time conversational recommender system that observes a conversation and proposes relevant advertisements. They apply it on conversations between users of online social networks. They train a Latent Dirichlet Allocation (LDA) model that performs topic modeling on the conversations and the advertisements, and propose the advertisement that is more similar by comparing the topic distributions. In contrast to our approach, their topics are of arbitrary meaning and do not have explainable properties though our approach performs user modeling on the directly interpretable items' categories.

\section{Methodology}
\label{sec:methodology}
\begin{figure}[t]
   \includegraphics[width=\textwidth]{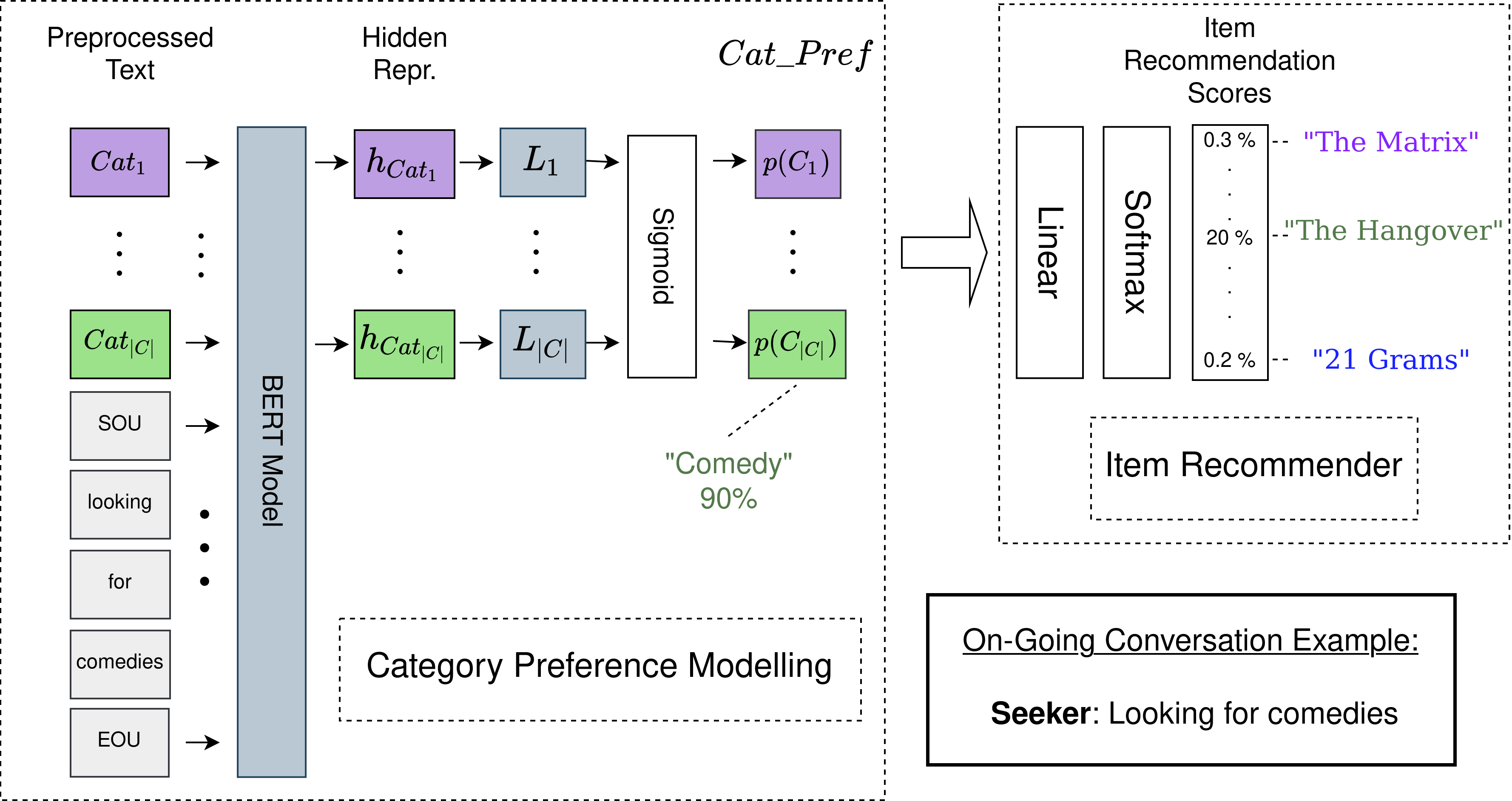}
\caption{The architecture of our proposed approach. In this example, the user initiates the conversation by saying "Looking for comedies". The Category Preference Modelling (left) predicts the user's interest for each category ($Cat\_Pref$) and the Item Recommender (right) translates the aforementioned prediction into a normalized distribution of item recommendation scores.} 
\label{fig:full_model}
\end{figure}

A conversational item recommendation system usually has to suggest several items, until user's criteria are met or she is given enough options.
Our method aims on replicating a recommendation behaviour, by using the recommended item as a target item (training signal) and the conversation history, at that moment, as input.
In this work we assume that each item falls into one or more categories (e.g. genre for movies, cuisine for recipes), and that user's criteria can take the form of category preferences.
Consequently, instead of directly predicting the target item, we firstly apply \textbf{category preference modelling} so to predict user's category preferences given the conversation history, and secondly the \textbf{item recommender} nominates items using the predicted categorical preferences.
Specifically, we first train the former module to predict the category vector of the target item, that we assume to best describe user's articulated categorical preferences by that moment. 
After this module is trained, we freeze its parameters and use its predictions as input in order to independently train the item recommendation module on recommending the target item.
This approach is intuitive and allows for the predicted user’s desired categories to be used for justifying the recommendation. 
In this section we thoroughly describe our method, a visual description of which is shown in Figure \ref{fig:full_model}.

\subsection{Category Preference Modelling}

This module is trained to predict the categories of the item that the system should suggest, given the earlier utterances of the conversation.
The training follows a supervised manner, having as target the category vector of the target item, $ [0,1]^{|C|} $ where $|C|$ is the number of categories.
Supervised training ensures that each dimension represents preference over some specific category.
During training we minimize Root Mean Square Error (RMSE).
The input of the module consists of the past utterances, and special tokens.
Each category is explicitly represented by an extra token at the beginning of the input ; $Cat_1, \dots, Cat_{|C|}$; as shown in Figure \ref{fig:full_model}. 
BERT generates a hidden representation for each of these tokens ($h_{Cat_1}, \dots, h_{Cat_{|C|}}$), while having as context the current history of the conversation, i.e. Eq. (\ref{eq:cat_pref_pred_hidden}). These hidden representations are being propagated by independent Linear Layers ($L_1, \dots, L_{|C|}$) followed by the Sigmoid activation function, i.e.~Eq. (\ref{eq:cat_pref_pred_sigmoid}). All category preference predictions are being concatenated, forming an output vector $Cat\_Pref \in [0,1]^{|C|}$ i.e.~ Eq. (\ref{eq:cat_pref_pred_final}).

\begin{equation}
\label{eq:cat_pref_pred_final}
Cat\_Pref^{|C|} = Concatenate( Cat_1\_Pref^1, \dots , Cat_{|C|}\_Pref^1) \in [0,1]^{|C|}
\end{equation}

\begin{equation}
\label{eq:cat_pref_pred_sigmoid}
Cat_i\_Pref^1 = Sigmoid(W_i^{1\times hid} \times h_{Cat_i}^{hid} + b_i^1)  \in [0,1]
\end{equation}

\begin{equation}
\label{eq:cat_pref_pred_hidden}
h_{Cat_i}^{hid} = BERT(Cat_i | History )  \in \mathbb{R}^{hid}
\end{equation}

Equations (\ref{eq:cat_pref_pred_final}) to (\ref{eq:cat_pref_pred_hidden}) describe the function of the Category Preference Modeling, where $\mathbb{R}$ denotes Real number and $hid$ is the hidden representation size of BERT. The BERT model is not part of our contribution. For further details regarding BERT's function please refer to the original paper \cite{BERT}.

\subsubsection*{Input Formation:}
The conversation history is forming an input that starts with the aforementioned $Cat_i$ tokens, followed by the utterances. The utterances are in their original chronological order, which are given at token level and each utterance is separated by the Separation \textit{SEP} token. The Start Of Utterance \textit{SOU} and End Of Utterance \textit{EOU} tokens are put at the beginning and the end of each utterance respectively. Furthermore, all items mentioned in the conversation are being replaced by an Item Mentioned (\textit{IM}) token and are being practically ignored. Finally, there are two types of segment embeddings; one for the tokens that originate form user's utterances and another for the system's tokens, including tokens that come from system's utterances and also the added special tokens ($Cat_i$, \textit{SEP}, \textit{SOU}, \textit{EOU}, \textit{IM}).

\subsection{Item Recommendation}
The second step of our approach is to translate category preferences to item recommendation scores. This is performed by a trainable Linear layer, followed by the Softmax activation function as described in Eq. (\ref{eq:cat2item}). This module is trained in a supervised manner, for the task of recommending the item that was suggested by the system at the given state of the conversation; minimizing Cross Entropy Loss.

\begin{equation}
\label{eq:cat2item}
Item\_Rec\_Scores^{|I|} = Softmax(W^{|I| \times |C|} \times Cat\_Pref^{|C|} + b^{|I|})
\end{equation}

Where $W$ and $b$ are trainable parameters, while $Cat\_Pref$ represents the prediction of our first step, as defined in Eq. (\ref{eq:cat_pref_pred_final}). Accordingly, their shape is defined by $|C|$ and $|I|$ which are the number of categories and items respectively.

\section{Experiments and Analysis}
In this section we describe the experiments that allow us to validate the recommendation performance of our proposed approach. 
The code of our experiments is publicly available, ensuring the reproducibility of our findings\footnote{https://github.com/kondilidisn/exp-conv-rec}. Regarding implementation details, we finetuned a pretrained BERT model for our Category Preference Modelling module, using the Transformers python library\footnote{https://github.com/huggingface/transformers}.

\subsubsection*{Dataset:}
We use the ReDial conversational recommendation dataset \cite{ReDial}, which consists of 10,006 real conversations including 51,699 total movie mentions, over 6924 unique movies. The conversations of the dataset are split to training (90\%) and test (10\%) sets, from which we further split the training set to training(72\%) and validation (18\%) sets. In each conversation there is a movie \textbf{Seeker} (user) and a movie \textbf{Recommender} (system). ReDial represents our desired problem setting very well, because each conversational recommendation session is considered individually, treating all users as new users (cold start setting), and it recommends movies that can be described by a common set of categories (genres). Additionally, at least four movies need to be mentioned in each conversation. For our approach, we retrieved categorical vectors for each item. Following \cite{ReDial}, we used the  MovieLens dataset\footnote{https://grouplens.org/datasets/movielens/latest/, generated in September 2018} \cite{MovieLens} for extracting categorical information of movies. There are 19 movie categories\footnote{`Comedy', `IMAX', `Romance', `Western', `Crime', `Sci-Fi', `Animation', `Thriller', `Fantasy', `Film-Noir', `Mystery', `Action', `Horror', `Adventure', `Musical', `Children', `Drama', `War', `Documentary'}, and one option for ``no category information''. As a result, a binary categorical vector is formed for each movie. A total of 1746 (25\%) of ReDial's movies were not successfully matched with any movie from MovieLens. For these movies, and for those that were characterized as ``no category information'', we set the category vector to be equal to 0.5 for each category. 

\subsubsection*{Experimental Details}
We follow the same experimental setup as earlier work on this dataset in order to measure item recommendation performance \cite{KBRD}.
The task at hand is to replicate the Recommender's (System's) behavior regarding item recommendation. 
Specifically, we assume all items mentioned by the Recommender to be correct recommendations, with respect to the conversation's history at that time \cite{ReDial}. 
Therefore, one sample is created for each item mentioned by the Recommender (target item) that needs to be predicted given the previous utterances of the conversation (input). As mentioned at the beginning of Sec \ref{sec:methodology}, this sample is split into two training samples; one for each module. The first module is trained on predicting the categorical vector of the mentioned item, given only the previous utterances, while the second module is then trained to identify the target item, given the categorical preference prediction. 
We evaluate recommendation performance using Recall at rank N (Rec@N) percentage (\%); N $\in \{1,10\}$ and average it over all target items. 
We use the validation set for hyper-parameter tuning and early stopping (5 epochs) and the test set for evaluating performance.

\subsubsection*{Compared Approaches:}
The presented experiments aim to position our model's item recommendation performance compared to similar literature and state-of-the-art approaches. Therefore, we use \textbf{ReDial} \cite{ReDial} and the current state-of-the-art in this dataset \textbf{KBRD} \cite{KBRD}. Compared to these baselines, our model has explainable properties, even though they are not being evaluated by online user studies. Furthermore, we test three models based on our approach. The first one, \textbf{Ours}, is our proposed approach as described in Section \ref{sec:methodology}. The second one, \textbf{Ours E2E}, shares the same architecture but the two modules are trained in an end-to-end setting, directly predicting the target item from the conversation and not taking advantage of the categorical information of the target items. Finally, as an oracle setting, \textbf{Ours + GT}, we assume the category preference prediction to be ideal, providing the true category vector of the target item to the second step of our approach, which is the only trained module in this setting.

\begin{table}[]
\caption{Comparing the performances of the compared models for the task of Conversational Item Recommendation task on the ReDial dataset.}
\label{tab:results}
\begin{tabularx}{\textwidth}{Xcc}
\hline
Model                & Rec@1 \% & Rec@10 \% \\ 
\hline
ReDial \cite{ReDial}               & 1.50  & 10.49  \\ 
KBRD \cite{KBRD}                 & 2.15  & \textbf{16.42}  \\ 
Ours E2E        & 0.95  & 5.98  \\ 
Ours                 & \textbf{2.37}  & 13.21 \\
\hline
Ours + GT & 31.85 & 62.73  \\
\hline
\end{tabularx}
\end{table}

\subsubsection*{How effective is our model's performance?}
The performance of the compared approaches are summarized in Table~\ref{tab:results}. Compared to the baselines, our proposed model \textbf{Ours} outperforms the ReDial, and achieve comparable results with KBRD. As \cite{rec_1_importance} suggests, Rec@1 is most important on conversational recommendation, since usually only one item is being suggested each time in such a setting. We recall, however, that the two baselines are not explainable.
Additionally, when training our model end-to-end (Ours E2E), the model is not explainable and its recommendation performance drops to less than half. This emphasizes the added value of utilizing categorical information in terms of recommendation performance. 
Finally, in the oracle setting (Ours + GT), where we assume a perfect category preference prediction, the performance is greatly improved which indicates that there is room for improvement in the category preference modelling of our approach, which will lead to improved recommendation results.

\subsubsection*{How effective is our explainability?}
\begin{figure}[h]
  \caption{Demonstrating accurate item recommendation and the model's ability to correctly predict preferred categories from keywords.}
  \label{fig:example}

\small
\begin{tabular*}{\textwidth}{l|l @{\extracolsep{\fill}}}
 \multicolumn{1}{c|}{\normalsize \textbf{Sender}} & \multicolumn{1}{c}{ \normalsize \textbf{Message} } \\ \hline
\multicolumn{1}{c|}{ \textbf{$\dots$}} & \multicolumn{1}{c}{  \textbf{$\dots$} } \\ \hline
Recommender \hspace*{0.6cm} & Great, what kinds of movies are you looking for ?\\\hline
\multirow{2}{*}{Seeker} &I 'm looking for Disney classics, like \textbf{Masked\_Item} \\ & to show my niece. Can you recommend any ?\\ \hline
\textbf{Ours} & \textbf{Applying Category Preference Modelling:} \\ 

\end{tabular*}

\begin{subfigure}{\textwidth}
\centering
\includegraphics[width=\textwidth]{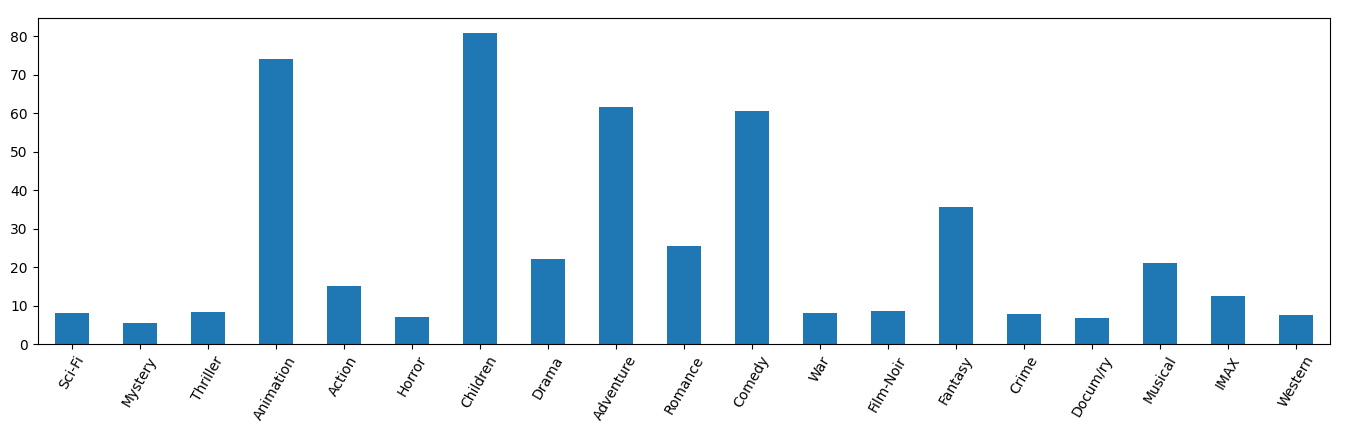}
\end{subfigure}

\small
\begin{tabular*}{\textwidth}{l|l @{\extracolsep{\fill}}}

\hline

\textbf{Our top 2 Rec.} & \textbf{Corresponding Items' Ground Truth Categories:} \\
\scriptsize 1. Moana (2016) &\scriptsize \{Animation, Children, Adventure, Comedy \& Fantasy \} \\
\scriptsize 2. Coco (2017) &\scriptsize \{Animation, Children \& Adventure \} \\
\textbf{Our Explanation} &  \textit{"Because you are looking for something that combines :"} \\
(Over 50\%) &\scriptsize Children(80\%), Animation(74\%), Adventure(61\%) \& Comedy(60\%) \\ 

\hline

\multirow{2}{*}{Recommender} & Recent films like \underline{Moana (2016)} and \underline{Zootopia} are \\ &  great. I also enjoyed Pixar's \underline{Coco  (2017)} \\\hline
\multirow{2}{*}{Seeker} & Hmm. I have not seen any of these. I am writing \\ & them down now! Thanks a lot!!\\\hline
  
\end{tabular*}

\end{figure}

\begin{figure}[h]

\caption{Demonstrating correct item-category mapping when this information is unavailable and accurate recommendation at the same time.}
\label{fig:example2}
\small
\begin{tabular*}{\textwidth}{l|l @{\extracolsep{\fill}}}
 \multicolumn{1}{c|}{\normalsize \textbf{Sender}} & \multicolumn{1}{c}{ \normalsize \textbf{Message} } \\ \hline
\multicolumn{1}{c|}{ \textbf{$\dots$}} & \multicolumn{1}{c}{  \textbf{$\dots$} } \\ \hline
\multirow{2}{*}{Seeker} \hspace*{1.7cm} & Do you know any good crime movies \\ & I really like crime movies\\ \hline
\textbf{Ours} & \textbf{Applying Category Preference Modelling:} \\ 

\end{tabular*}

\begin{subfigure}{\textwidth}
\centering
\includegraphics[width=\textwidth]{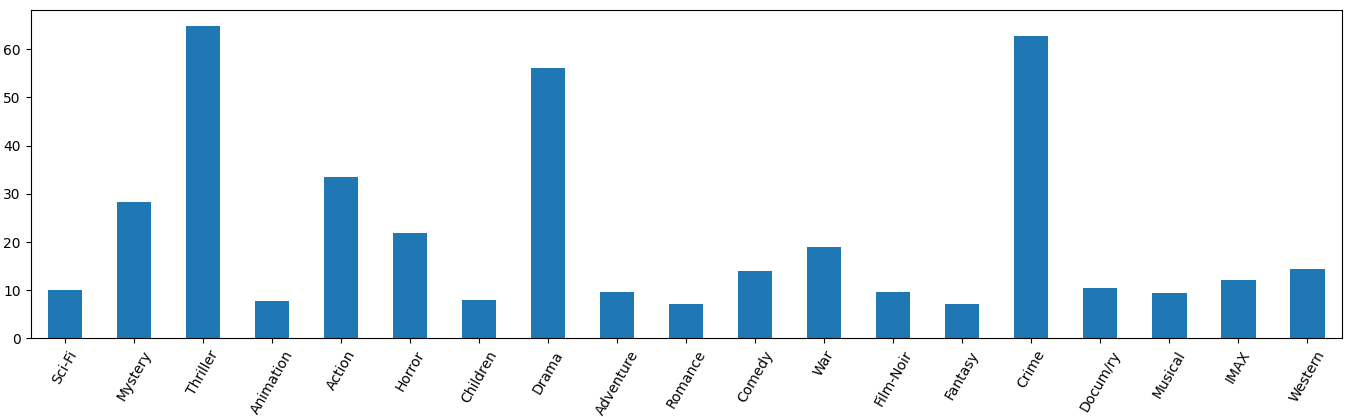}
\end{subfigure}

\small
\begin{tabular*}{\textwidth}{l|l @{\extracolsep{\fill}}}

\hline

\textbf{Our top 2 Rec.} &\textbf{Corresponding Items' Ground Truth Categories:} \\ 
\scriptsize  1. Seven (1995) &\scriptsize  \textbf{No Categorical Information Available} \\
\scriptsize  2. Zodiac (2007) &\scriptsize  \{Thriller, Drama \& Crime \} \\ 
\textbf{Our Explanation} &  \textit{"Because you are looking for something that combines :"} \\
(Over 50\%) &\scriptsize  Thriller(64\%), Crime(62\%) \& Drama(56\%) \\ \hline

Recommender & Yes, I love \underline{Seven (1995)} and \underline{Godfather (1991)} \\\hline

\end{tabular*}
\end{figure}

As \cite{viewpoint_regr} points out, it is common to present examples for assessing the explainability of a model. To that end, we demonstrate our two-step approach on two conversations of ReDial's test set in Figures \ref{fig:example} and \ref{fig:example2}. 

On the example presented in Figure \ref{fig:example}, even though no category is explicitly mentioned, our proposed category preference modelling approach is able to translate ``Disney Classics'' and ``to show my niece'' into the preference for the categories `Children', `Animation', `Adventure' and `Comedy', in order of importance. Moreover, our item recommender is able to use our category preference prediction and recommend two out of the three movies that the Recommender is about to suggest. Finally, the predicted preferred categories are overlapping with the ground truth categories of the target items with high accuracy. 

On the second example shown in Figure \ref{fig:example2}, our model is able to correctly predict the movie that the Recommender is about to suggest. Additionally, the correct recommendation is due to the three predicted preferred categories \{Thriller, Crime, Drama \} that include the category that the user mentions "crime movies". This demonstrates the potential ability of our model to correlate items with categories, when this information is unavailable. It should be reminded that 25\% of the items used in our experiments are lacking categorical information making this ability very useful.

\section{Conclusion \& Future Work}

In this work, we propose a novel explainable recommender on the conversational setting that can be applied in real time and in a cold-start setting.
Furthermore, the performance of our approach is comparable to the state-of-the-art which is not explainable while our model is more intuitive and provides very concise explanations in the form of category preferences. We also show that there is room for improvement regarding the recommendation performance of our proposed model where the future work should focus on. In addition, future studies can take advantage of the explainable nature of our user modelling, and directly incorporate user feedback on the explanation level (i.e. categories in our work), since more sophisticated feedback of that type can lead to reducing the number of interactions needed for effective recommendation to 40\% \cite{interactive_feedback}. We focus on improving the recommendation performance for conversational recommender, and thus no dialogue generation model is incorporated. One can integrate dialogue generation with the recommended items embedded. Finally, we leave a user survey to qualitatively and quantitatively evaluate the explanations provided to future work.

\section{Acknowledgments}
This research was supported by
the NWO Innovational Research Incentives Scheme Vidi (016.Vidi.189.039),
the NWO Smart Culture - Big Data / Digital Humanities (314-99-301),
the H2020-EU.3.4. - SOCIETAL CHALLENGES - Smart, Green And Integrated Transport (814961)
All content represents the opinion of the authors, which is not necessarily shared or endorsed by their respective employers and/or sponsors.

\begin{figure}[b]
    \centering
    \includegraphics[width=\textwidth]{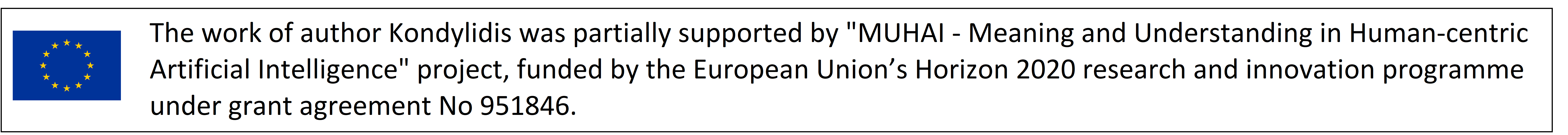}
\end{figure}

\clearpage

\bibliographystyle{splncs04}
\bibliography{main}

\begin{thebibliography}{10}
\providecommand{\url}[1]{\texttt{#1}}
\providecommand{\urlprefix}{URL }
\providecommand{\doi}[1]{https://doi.org/#1}

\bibitem{real_time_conv_rec}
Albalawi, R., Yeap, T.H., Benyoucef, M.: Toward a real-time social
  recommendation system. In: Proceedings of the 11th International Conference
  on Management of Digital EcoSystems. p. 336–340. MEDES '19, Association for
  Computing Machinery, New York, NY, USA (2019). \doi{10.1145/3297662.3365789},
  \url{https://doi.org/10.1145/3297662.3365789}

\bibitem{real_time_social_rec}
Albalawi, R., Yeap, T.H., Benyoucef, M.: Toward a real-time social
  recommendation system. In: Proceedings of the 11th International Conference
  on Management of Digital EcoSystems. p. 336–340. MEDES ’19, Association
  for Computing Machinery, New York, NY, USA (2019).
  \doi{10.1145/3297662.3365789}, \url{https://doi.org/10.1145/3297662.3365789}

\bibitem{KBRD}
Chen, Q., Lin, J., Zhang, Y., Ding, M., Cen, Y., Yang, H., Tang, J.: Towards
  knowledge-based recommender dialog system (2019)

\bibitem{expl_conv_rec}
Chen, Z., Wang, X., Xie, X., Parsana, M., Soni, A., Ao, X., Chen, E.: Towards
  explainable conversational recommendation. In: International Joint
  Conferences on Artificial Intelligence (IJCAI) (May 2020),
  \url{https://www.microsoft.com/en-us/research/publication/towards-explainable-conversational-recommendation/}

\bibitem{MMALFM}
Cheng, Z., Chang, X., Zhu, L., {Catherine Kanjirathinkal}, R., Kankanhalli,
  M.S.: {MMALFM:} explainable recommendation by leveraging reviews and images.
  CoRR  \textbf{abs/1811.05318} (2018), \url{http://arxiv.org/abs/1811.05318}

\bibitem{Christakopoulou}
Christakopoulou, K., Radlinski, F., Hofmann, K.: Towards conversational
  recommender systems. In: Proceedings of the 22nd ACM SIGKDD International
  Conference on Knowledge Discovery and Data Mining. p. 815–824. KDD '16,
  Association for Computing Machinery, New York, NY, USA (2016).
  \doi{10.1145/2939672.2939746}, \url{https://doi.org/10.1145/2939672.2939746}

\bibitem{KBRD_entity_linking}
Daiber, J., Jakob, M., Hokamp, C., Mendes, P.N.: Improving efficiency and
  accuracy in multilingual entity extraction. In: Proceedings of the 9th
  International Conference on Semantic Systems. p. 121–124. I-SEMANTICS
  ’13, Association for Computing Machinery, New York, NY, USA (2013).
  \doi{10.1145/2506182.2506198}, \url{https://doi.org/10.1145/2506182.2506198}

\bibitem{BERT}
Devlin, J., Chang, M., Lee, K., Toutanova, K.: {BERT:} pre-training of deep
  bidirectional transformers for language understanding. CoRR
  \textbf{abs/1810.04805} (2018), \url{http://arxiv.org/abs/1810.04805}

\bibitem{Switch}
G{\"{u}}l{\c{c}}ehre, {\c{C}}., Ahn, S., Nallapati, R., Zhou, B., Bengio, Y.:
  Pointing the unknown words. CoRR  \textbf{abs/1603.08148} (2016),
  \url{http://arxiv.org/abs/1603.08148}

\bibitem{MovieLens}
Harper, F.M., Konstan, J.A.: The movielens datasets: History and context. ACM
  Trans. Interact. Intell. Syst.  \textbf{5}(4),  19:1--19:19 (Dec 2015).
  \doi{10.1145/2827872}, \url{http://doi.acm.org/10.1145/2827872}

\bibitem{dbpedia}
Lehmann, J., Isele, R., Jakob, M., Jentzsch, A., Kontokostas, D., Mendes, P.N.,
  Hellmann, S., Morsey, M., van Kleef, P., Auer, S., Bizer, C.: {DBpedia} - a
  large-scale, multilingual knowledge base extracted from wikipedia. Semantic
  Web Journal  \textbf{6}(2),  167--195 (2015),
  \url{http://jens-lehmann.org/files/2015/swj\_dbpedia.pdf}

\bibitem{ReDial}
Li, R., Ebrahimi~Kahou, S., Schulz, H., Michalski, V., Charlin, L., Pal, C.:
  Towards deep conversational recommendations. In: Bengio, S., Wallach, H.,
  Larochelle, H., Grauman, K., Cesa-Bianchi, N., Garnett, R. (eds.) Advances in
  Neural Information Processing Systems 31, pp. 9725--9735. Curran Associates,
  Inc. (2018),
  \url{http://papers.nips.cc/paper/8180-towards-deep-conversational-recommendations.pdf}

\bibitem{conv_travel}
Liao, L., Takanobu, R., Ma, Y., Yang, X., Huang, M., Chua, T.: Deep
  conversational recommender in travel. CoRR  \textbf{abs/1907.00710} (2019),
  \url{http://arxiv.org/abs/1907.00710}

\bibitem{rec_1_importance}
Liao, L., Takanobu, R., Ma, Y., Yang, X., Huang, M., Chua, T.: Deep
  conversational recommender in travel. CoRR  \textbf{abs/1907.00710} (2019),
  \url{http://arxiv.org/abs/1907.00710}

\bibitem{fashion_comment_generation}
Lin, Y., Ren, P., Chen, Z., Ren, Z., Ma, J., de~Rijke, M.: Explainable fashion
  recommendation with joint outfit matching and comment generation. CoRR
  \textbf{abs/1806.08977} (2018), \url{http://arxiv.org/abs/1806.08977}

\bibitem{visual_fusion}
Liu, P., Zhang, L., Gulla, J.A.: Dynamic attention-based explainable
  recommendation with textual and visual fusion. Information Processing \&
  Management p. 102099 (2019). \doi{https://doi.org/10.1016/j.ipm.2019.102099},
  \url{http://www.sciencedirect.com/science/article/pii/S0306457319301761}

\bibitem{aspera}
Nikolenko, S.I., Tutubalina, E., Malykh, V., Shenbin, I., Alekseev, A.: Aspera:
  Aspect-based rating prediction model. CoRR  \textbf{abs/1901.07829} (2019),
  \url{http://arxiv.org/abs/1901.07829}

\bibitem{conv_rec_importance_1}
Radlinski, F., Balog, K., Byrne, B., Krishnamoorthi, K.: Coached conversational
  preference elicitation: A case study in understanding movie preferences. pp.
  353--360 (01 2019). \doi{10.18653/v1/W19-5941}

\bibitem{viewpoint_regr}
Ren, Z., Liang, S., Li, P., Wang, S., Rijke, M.: Social collaborative viewpoint
  regression with explainable recommendations. pp. 485--494 (02 2017).
  \doi{10.1145/3018661.3018686}

\bibitem{towards_graph}
Samih, A., Adadi, A., Berrada, M.: Towards a knowledge based explainable
  recommender systems. pp.~1--5 (10 2019). \doi{10.1145/3372938.3372959}

\bibitem{autorec}
Sedhain, S., Menon, A.K., Sanner, S., Xie, L.: Autorec: Autoencoders meet
  collaborative filtering. In: Proceedings of the 24th International Conference
  on World Wide Web. pp. 111--112. WWW '15 Companion, ACM, New York, NY, USA
  (2015). \doi{10.1145/2740908.2742726},
  \url{http://doi.acm.org/10.1145/2740908.2742726}

\bibitem{HRED}
Sordoni, A., Bengio, Y., Vahabi, H., Lioma, C., Simonsen, J.G., Nie, J.: A
  hierarchical recurrent encoder-decoder for generative context-aware query
  suggestion. CoRR  \textbf{abs/1507.02221} (2015),
  \url{http://arxiv.org/abs/1507.02221}

\bibitem{pretrained_gensen_representations}
Subramanian, S., Trischler, A., Bengio, Y., Pal, C.J.: Learning general purpose
  distributed sentence representations via large scale multi-task learning.
  CoRR  \textbf{abs/1804.00079} (2018), \url{http://arxiv.org/abs/1804.00079}

\bibitem{CRS}
Sun, Y., Zhang, Y.: Conversational recommender system. CoRR
  \textbf{abs/1806.03277} (2018), \url{http://arxiv.org/abs/1806.03277}

\bibitem{DARIA_SARAH}
{Tal}, O., {Liu}, Y., {Huang}, J., {Yu}, X., {Aljbawi}, B.: Neural attention
  frameworks for explainable recommendation. IEEE Transactions on Knowledge and
  Data Engineering pp.~1--1 (2019). \doi{10.1109/TKDE.2019.2953157}

\bibitem{the_FacT}
Tao, Y., Jia, Y., Wang, N., Wang, H.: The fact: Taming latent factor models for
  explainability with factorization trees. CoRR  \textbf{abs/1906.02037}
  (2019), \url{http://arxiv.org/abs/1906.02037}

\bibitem{transformers}
Vaswani, A., Shazeer, N., Parmar, N., Uszkoreit, J., Jones, L., Gomez, A.N.,
  Kaiser, L., Polosukhin, I.: Attention is all you need (2017)

\bibitem{openionated}
Wang, N., Wang, H., Jia, Y., Yin, Y.: Explainable recommendation via multi-task
  learning in opinionated text data. In: The 41st International ACM SIGIR
  Conference on Research \&\#38; Development in Information Retrieval. pp.
  165--174. SIGIR '18, ACM, New York, NY, USA (2018).
  \doi{10.1145/3209978.3210010},
  \url{http://doi.acm.org/10.1145/3209978.3210010}

\bibitem{interactive_feedback}
Yu, T., Shen, Y., Jin, H.: A visual dialog augmented interactive recommender
  system. In: Proceedings of the 25th ACM SIGKDD International Conference on
  Knowledge Discovery \& Data Mining. p. 157–165. KDD ’19, Association for
  Computing Machinery, New York, NY, USA (2019). \doi{10.1145/3292500.3330991},
  \url{https://doi.org/10.1145/3292500.3330991}

\bibitem{SAUR}
Zhang, Y., Chen, X., Ai, Q., Yang, L., Croft, W.B.: Towards conversational
  search and recommendation: System ask, user respond. In: Proceedings of the
  27th ACM International Conference on Information and Knowledge Management.
  pp. 177--186. CIKM '18, ACM, New York, NY, USA (2018).
  \doi{10.1145/3269206.3271776},
  \url{http://doi.acm.org/10.1145/3269206.3271776}

\bibitem{explicit_factor_models}
Zhang, Y., Lai, G., Zhang, M., Zhang, Y., Liu, Y., Ma, S.: Explicit factor
  models for explainable recommendation based on phrase-level sentiment
  analysis. In: Proceedings of the 37th International ACM SIGIR Conference on
  Research \& Development in Information Retrieval. p. 83–92. SIGIR ’14,
  Association for Computing Machinery, New York, NY, USA (2014).
  \doi{10.1145/2600428.2609579}, \url{https://doi.org/10.1145/2600428.2609579}

\bibitem{EXPLORE}
Zheng, X., Wang, M., Chen, C., Wang, Y., Cheng, Z.: Explore: Explainable
  item-tag co-recommendation. Information Sciences  \textbf{474},  170 -- 186
  (2019). \doi{https://doi.org/10.1016/j.ins.2018.09.054},
  \url{http://www.sciencedirect.com/science/article/pii/S0020025518307667}

\bibitem{zou2020towardsb}
Zou, J., Chen, Y., Kanoulas, E.: Towards question-based recommender systems.
  In: Proceedings of the 43rd International ACM SIGIR Conference on Research
  and Development in Information Retrieval. pp. 881--890. SIGIR '20 (2020)

\bibitem{zou2019learning}
Zou, J., Kanoulas, E.: Learning to ask: Question-based sequential bayesian
  product search. In: Proceedings of the 28th ACM International Conference on
  Information and Knowledge Management. pp. 369--378. CIKM '19 (2019)

\bibitem{zou2020towards}
Zou, J., Kanoulas, E.: Towards question-based high-recall information
  retrieval: Locating the last few relevant documents for technology-assisted
  reviews. ACM Trans. Inf. Syst.  \textbf{38}(3) (May 2020)

\bibitem{zou2018technology}
Zou, J., Li, D., Kanoulas, E.: Technology assisted reviews: Finding the last
  few relevant documents by asking yes/no questions to reviewers. In: The 41st
  International ACM SIGIR Conference on Research \& Development in Information
  Retrieval. pp. 949--952. SIGIR '18 (2018)

\end{thebibliography}

\end{document}